\documentclass[runningheads]{llncs}

\usepackage{eccv}

\usepackage{eccvabbrv}

\usepackage{graphicx}
\usepackage{booktabs}
\usepackage{multirow}
\usepackage{colortbl}
\newcommand{\darkgreen}[1]{\textcolor[rgb]{0.00,0.70,0.00}{#1}}

\usepackage[accsupp]{axessibility}  

\definecolor{MyCiteColor}{rgb}{0, 0.7, 1}
\usepackage[pagebackref,colorlinks,citecolor=MyCiteColor]{hyperref}
\usepackage{orcidlink}

\begin{document}

\title{VTAgent: Agentic Keyframe Anchoring for Evidence-Aware Video TextVQA} 

\titlerunning{VTAgent}

\author{Haibin He\inst{1}\orcidlink{0009-0001-2004-150X} \and
Maoyuan Ye\inst{1}\orcidlink{0000-0002-4180-1096} \and
Jing Zhang\inst{1}$^{\dagger}$\orcidlink{0000-0001-6595-7661} \and
Juhua Liu\inst{1}$^{\dagger}$\orcidlink{0000-0002-3907-8820} \and
Bo Du\inst{1}\orcidlink{0000-0002-0059-8458}}

\authorrunning{He et al.}

\institute{School of Computer Science, National Engineering Research Center for Multimedia Software, Institute of Artificial Intelligence, and Hubei Key Laboratory of Multimedia and Network Communication Engineering, Wuhan University, Wuhan, Hubei, China 
\email{\{haibinhe,yemaoyuan,liujuhua,dubo\}@whu.edu.cn}, jingzhang.cv@gmail.com \\
${\dagger}$ Corresponding author}

\maketitle

\begin{abstract}
  Video text-based visual question answering (Video TextVQA) aims to answer questions by reasoning over visual textual content appearing in videos. Despite the strong multimodal video understanding capabilities of recent Video-LLMs, their performance on existing Video TextVQA benchmarks remains limited. To better understand this gap, we conduct an upper-bound analysis through frame-wise question answering, counting a sample as correct if any frame yields the right answer, which significantly outperforms direct video-based inference and reveals a substantial performance gap. The results suggest that the primary bottleneck lies in the localization of key question-relevant evidence, rather than in reasoning capacity itself.
Building on this insight, we propose a question-guided agent framework that explicitly anchors the relevant keyframes before answering. The approach operates effectively in a training-free setting and consistently surpasses direct video inference. With additional supervised fine-tuning (SFT) and reinforcement learning (RL), it achieves an average improvement of $+12.12$ in accuracy and $+11.15$ in ANLS across benchmarks, establishing new state-of-the-art results.
Our study underscores the critical role of explicit keyframe anchoring for advancing Video TextVQA. 
The code will be publicly released.
  \keywords{Video TextVQA \and Keyframe Anchoring \and Agent Learning}
\end{abstract}

\section{Introduction}
\label{sec:intro}
Understanding visual textual information in videos is fundamental to advancing comprehensive video intelligence~\cite{shi2025mme}. Beyond serving as auxiliary visual cues, scene text often conveys explicit semantic content not inferable from visual appearance alone. Accurate perception and interpretation of such textual content are therefore critical for robust semantic understanding and support various practical applications, including video understanding~\cite{zhao2023vtlayout}, video retrieval~\cite{sanders2023multivent,wu2025large} and autonomous driving~\cite{zhang2021character,zablocki2022explainability}.
To foster systematic progress in this direction, the research community has introduced the Video TextVQA task, along with dedicated benchmarks~\cite{zhao2022towards, tom2023reading}, to evaluate and promote models’ capabilities in perceiving and reasoning over scene text within video. Unlike static TextVQA, this task demands joint modeling of dynamic visual scenes, temporally distributed text, and vision-language understanding. Moreover, the textual evidence relevant to a question may appear only in a few frames, rendering precise evidence localization both challenging and essential for accurate reasoning.

Current approaches~\cite{zhao2022towards,zhang2025track,zhang2025gather,yan2026tom} tailored for the Video TextVQA task typically adopted a two-stage paradigm. These methods first employed external visual encoders(\eg,~\cite{ren2015faster,wang2023all}) and OCR systems (\eg,~\cite{ye2024hi,he2024gomatching}) to extract textual and visual features from video frames, which were subsequently fused for answer prediction. Benefiting from specialized OCR tools, such frameworks demonstrated strong text recognition capability and outperformed contemporary Video-LLMs~\cite{lin2024video,cheng2024videollama,wang2024qwen2,liu2025nvila} at the time. With the rapid evolution of large-scale multimodal learning, however, modern Video-LLMs have substantially improved their inherent OCR and reasoning capacities through expanded visual text-rich training data, refined optimization strategies, and architectural innovations. Consequently, general-purpose Video-LLMs now markedly surpass earlier small-scale, task-specific Video TextVQA models. Despite this progress, their performance remains still limited, indicating that improved OCR and reasoning abilities alone do not fully resolve the challenge.

Recent efforts have explored adapting Video-LLMs to Video TextVQA. For example, SFA~\cite{he2025sfa} introduces a video text detector~\cite{he2025gomatching++} to localize candidate regions containing visual text, followed by a scoring model that evaluates their relevance with the question and retains the most pertinent regions, thereby guiding the reasoning model to focus on critical textual evidence and improving the performance. Nevertheless, its performance ceiling is inherently constrained by the reliability of the external tools and susceptible to error accumulation.
These limitations prompt a more fundamental inquiry: \textit{what truly restricts the performance of current Video-LLMs on the Video TextVQA task}?

\begin{figure*}[t]
\centering
  \subfloat[Comparison of video-level and frame-wise inference on validation sets of M4-ViteVQA.]{\includegraphics[width=0.49\textwidth]{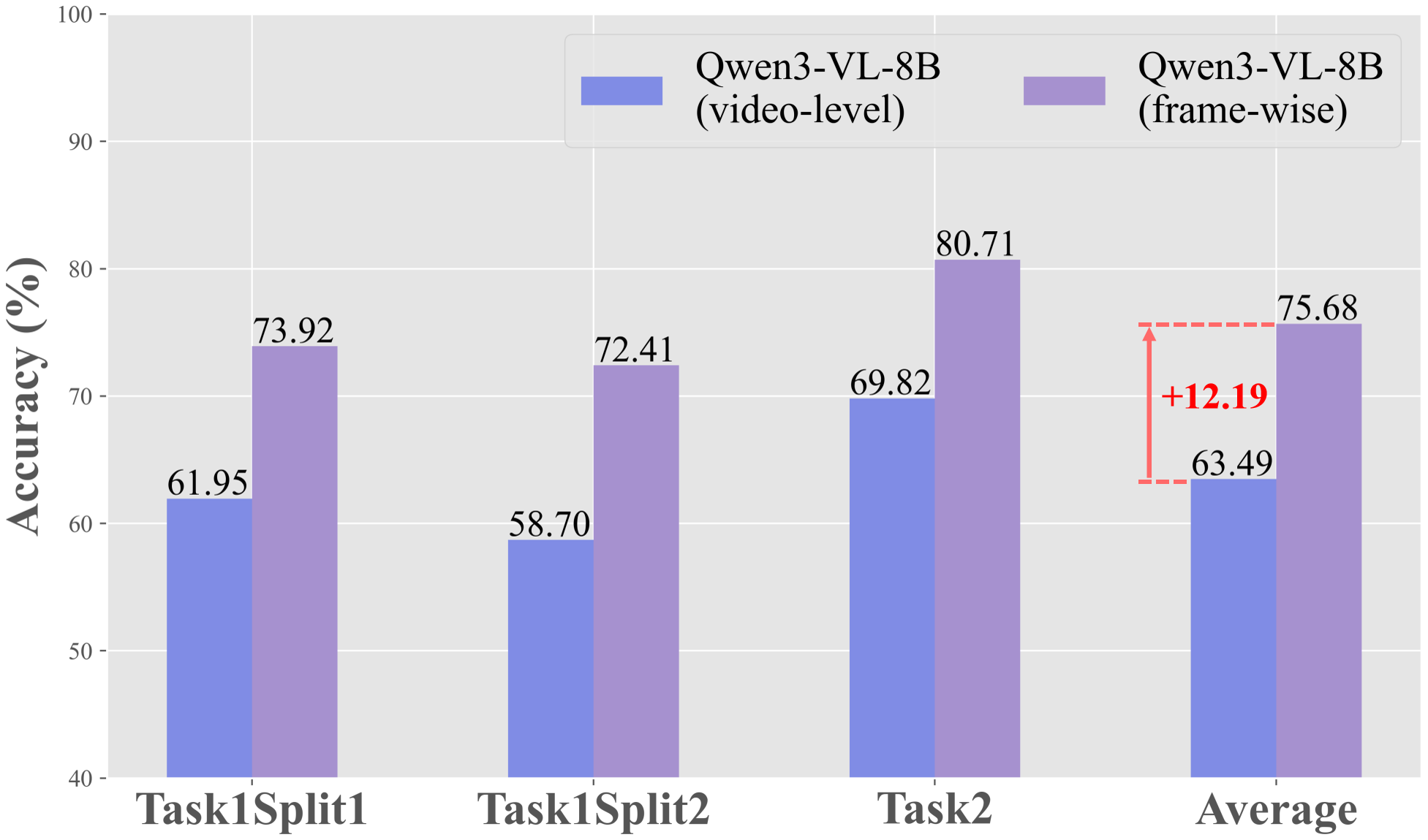}}
  \hfill
  \subfloat[Overall performance of VTAgent versus existing methods.]{\includegraphics[width=0.49\textwidth]{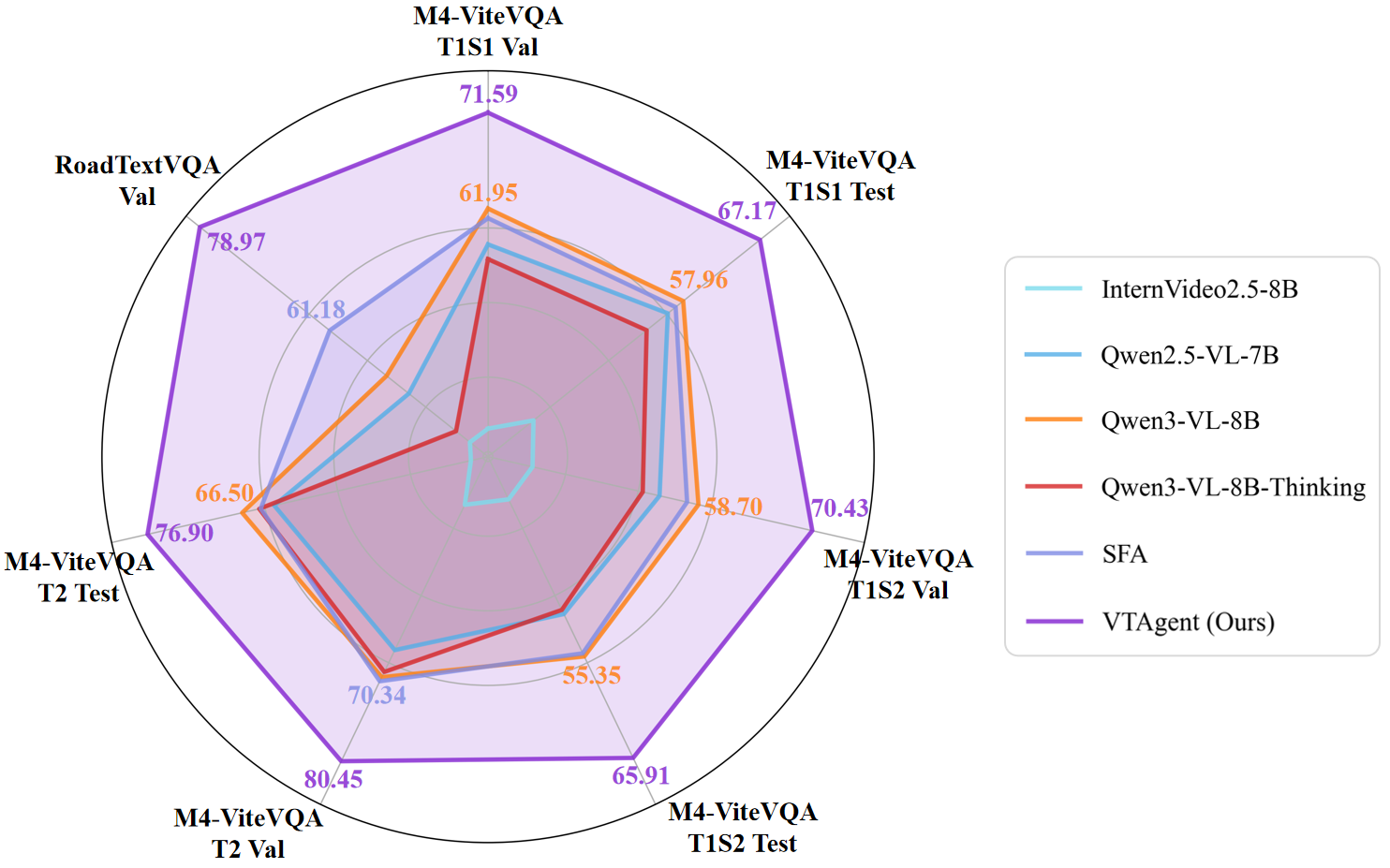}}
  \caption{\textbf{Illustration of the key motivation and effectiveness of VTAgent.} (a) shows the performance gap between frame-wise and holistic video inference, emphasizing the importance of accurately localizing question-relevant frames; (b) demonstrates VTAgent’s superior performance compared to existing approaches, validating the effectiveness of its locate-and-focus agentic framework.}
  \vspace{-5pt}
 \label{fig:pre}
\end{figure*}

To investigate this issue, we conduct a simple yet revealing upper-bound analysis. Instead of performing inference on the video input, we evaluate frame-wise question answering and regard a question as correctly solved if any individual frame yields the right answer. This oracle evaluation significantly outperforms direct video-level inference, exposing a pronounced performance gap. Specifically, as illustrated in Fig.~\ref{fig:pre} (a), frame-wise answering consistently outperforms video-level inference across all validation sets of the sub-tasks in M4-ViteVQA. On average, this setting achieves improvements of $+12.19$ in accuracy. Such a substantial margin indicates that current Video-LLMs are often capable of generating correct answers once the evidence-relevant frame is properly identified. In contrast, when required to jointly process the full temporal sequence, excessive redundancy and motion-induced text blurring or distortion may obscure critical information, making reliable evidence extraction more difficult, thereby leading to performance degradation.
These findings suggest that the dominant bottleneck of current Video-LLMs does not stem from their reasoning capacity per se, but rather from their inability to accurately localize question-relevant textual evidence within the video.

Motivated by this insight, we propose VTAgent, an agentic locate-and-focus framework for Video TextVQA that models the task as a sequential decision-making process. Rather than performing holistic video reasoning, VTAgent explicitly decomposes inference into two ordered actions: keyframe anchoring and keyframe-conditioned reasoning.
In the first stage, the agent actively analyzes the video frames conditioned on the question and selects keyframes that are most likely to contain relevant textual evidence. This anchoring step functions as an evidence-seeking action, filtering redundant or noisy frames through question-guided selection. In the second stage, the agent performs fine-grained reasoning over the selected keyframes to generate the final answer, conditioning its prediction on the explicitly localized evidence.
By structuring the task in this locate-and-focus paradigm, VTAgent directly addresses the evidence localization bottleneck revealed by our analysis, guiding the model to focus on critical visual content while reducing interference from temporal redundancy and motion-induced text degradation.
Notably, the proposed agentic framework can operate effectively in a training-free setting and consistently delivers stronger performance than direct inference with the base model. Furthermore, incorporating supervised fine-tuning (SFT) and reinforcement learning (RL) achieves additional performance gains, yielding an average improvement of $+12.12$ in accuracy and $+11.15$ in ANLS across benchmarks. These results establish new state-of-the-art performance while substantiating the effectiveness of VTAgent in explicit evidence localization, as shown in Fig.~\ref{fig:pre} (b).

In summary, our contributions are three-fold:

\quad $\cdot$ We conduct an oracle upper-bound analysis that identifies evidence localization as the primary bottleneck in Video TextVQA.

\quad $\cdot$ We propose VTAgent, an agentic locate-and-focus framework via two steps: keyframe anchoring and keyframe-conditioned reasoning, enabling explicit evidence localization and grounded answer generation.

\quad $\cdot$ Comprehensive evaluations across benchmarks validate the effectiveness of VTAgent, yielding consistent gains in both training-free and fine-tuned settings and advancing the state-of-the-art in Video TextVQA.

\section{Related Work}
\subsection{Video TextVQA}
Current mainstream task-specific approaches to Video TextVQA integrate OCR systems with multimodal feature fusion strategies to jointly model textual and visual cues in temporally evolving video content.
For instance, T5-ViteVQA~\cite{zhao2022towards} adopts multiple transformers for encoding OCR tokens, questions, and video representations, which are subsequently integrated by a multimodal fusion transformer to facilitate final answer generation.
TEA~\cite{zhang2025track} employs a complementary strategy to reconstruct the spatiotemporal dependencies of both scene text and objects, while simultaneously leveraging OCR-aware features to filter out unrelated information and enhance reasoning fidelity. 
GAT~\cite{zhang2025gather} further promotes the performance by coupling an off-the-shelf video text spotting (VTS) model~\cite{he2024gomatching, he2025gomatching++} with an instance-focused trajectory tracing module, effectively capturing the spatiotemporal dynamics for robust reasoning. 
More recently, breaking away from static OCR dependencies, TOM~\cite{yan2026tom} develops a unified framework featuring a trainable question-aware text spotter~\cite{ye2023deepsolo} refined by reasoning feedback, demonstrating strong generalization to Video TextVQA. 
In contrast, SFA~\cite{he2025sfa} largely enhances the performance by applying a training-free visual-cropping mechanism on Video Large Language Models (LLMs)~\cite{bai2025qwen2}.

\subsection{Video-LLMs}
Recent strides in LLMs~\cite{grattafiori2024llama3,cai2024internlm2,yang2025qwen3} have propelled the advancement of Video-LLMs which leverage LLMs as the foundational interface to bolster comprehensive video understanding and reasoning.
Particularly, founded on InternViT~\cite{chen2024expanding} and InternLM2.5~\cite{cai2024internlm2}, InternVideo2.5~\cite{wang2025internvideo2} augments the capabilities in high-fidelity visual perception and long-horizon temporal reasoning by harnessing Long and Rich Context Modeling. 
To efficiently process high-resolution images and long videos, NVILA~\cite{liu2025nvila} first scales up the spatial and temporal resolutions, and then compresses the visual tokens. 
By synergizing dynamic-resolution processing with absolute time encoding and a native resolution ViT, Qwen2.5-VL~\cite{bai2025qwen2} facilitates accurate long-document and long-video comprehension. 
Subsequently, Qwen3-VL~\cite{Qwen3-VL} introduces an enhanced interleaved-MRoPE for robust cross-modal spatial-temporal representation and devises an explicit text-based time alignment mechanism to achieve granular temporal grounding.
Albeit the promising capabilities exhibited by contemporary Video-LLMs, we conduct a preliminary experiment and reveal that the primary bottleneck for solving existing Video TextVQA problems lies in the localization of key-frames with question-relevant evidence. 
Consequently, we investigate a question-guided agent framework that explicitly anchors the key frame prior to final answering, effectively suppressing interference from redundant visual information.

\section{Method}
\label{sec:method}
To address the evidence localization bottleneck of current Video-LLMs on Video TextVQA task, we propose \textbf{VTAgent}, an agentic locate-and-focus framework for Video TextVQA that decomposes the task into two sequential stages: keyframe anchoring and keyframe-conditioned reasoning, as detailed in Section~\ref{sec:3.1}. Furthermore, to fully elicit the agentic capabilities of MLLMs, we design a systematic training pipeline that progressively enhances the model through supervised fine-tuning and reinforcement learning. The training procedure is described in Section~\ref{sec:3.2} and Section~\ref{sec:3.3}.

\subsection{VTAgent Framework}
\label{sec:3.1}
\begin{figure}[t!]
  \centering
    \includegraphics[width=\textwidth]{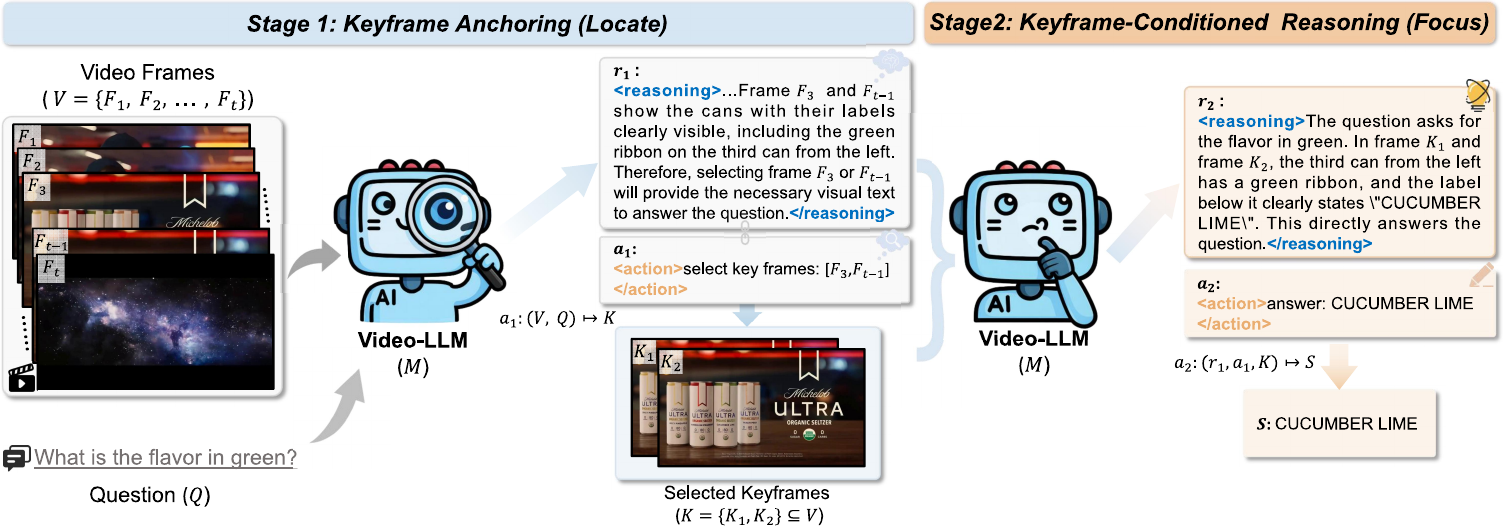}
    \caption{\textbf{Overview of VTAgent.} VTAgent performs keyframe anchoring and keyframe-conditioned reasoning to generate reliable answers from identified keyframes.}
    \vspace{-5pt}
\label{fig:framework}
\end{figure}

As illustrated in Fig.~\ref{fig:framework}, VTAgent is a structured framework for Video TextVQA that performs two-turn reasoning in an agentic manner. The framework consists of two sequential actions: first, the model analyzes the provided video frames together with the question and identifies keyframes that contain visual textual information relevant to answering the question; second, the model performs a detailed analysis of the selected keyframes and generates the final answer. By explicitly decomposing the task in this locate-and-focus paradigm, VTAgent guides the model to focus on question-relevant visual content, reducing the interference from redundant or noisy frames while providing an interpretable intermediate representation.

Let $V = \{F_1, F_2, \dots, F_t \}$ and $Q$ denote the input video frames and the associated question. We model the reasoning procedure of VTAgent as a two-step decision process with an action space 
$\mathcal{A} = \{a_1, a_2\}$, where $a_1$ corresponds to keyframe anchoring and $a_2$ corresponds to generating the final answer $S$.
The selected keyframes form a subset $K = \{K_1, K_2, \dots, K_n\} \subseteq V$.

In the first step, the model $M$ analyzes the entire $V$ in conjunction with the $Q$ and identifies a subset of question-relevant keyframes $K$. This keyframe anchoring stage can be formulated as:
\begin{equation}
    (r_1, a_1) = M(V, Q),
    \label{eq:1}
\end{equation}
where $r_1$ denotes the intermediate reasoning trace produced by the model to support keyframe extraction.

The corresponding action $a_1$ can be equivalently expressed as the following mapping:
\begin{equation}
    a_1 : (V, Q) \mapsto K,
    \label{eq:2}
\end{equation}

Conditioned on the anchored keyframes $K$, the model then performs keyframe-conditioned reasoning to generate the final answer.
This answer generation stage is defined as:
\begin{equation}
    (r_2, a_2) = M(r_1, a_1, K),
    \label{eq:3}
\end{equation}
where $r_2$ represents the reasoning process over the visual and textual content of the selected keyframes.

The answer action $a_2$ can be expressed as follows:
\begin{equation}
    a_2 : (r_1, a_1, K) \mapsto S.
    \label{eq:4}
\end{equation}

Overall, the complete reasoning trajectory $\tau$ for the question $Q$ can be represented as:
\begin{equation}
    \tau = \big((r_1, a_1), (r_2, a_2)\big),
    \label{eq:5}
\end{equation}
which captures the sequential process of keyframe anchoring followed by keyframe-conditioned reasoning.

To ensure structural consistency and controllability, each trace adheres to a predefined format, with intermediate reasoning and decisions encapsulated by the tags $<reasoning>\dots<reasoning>$ and $<action>\dots<action>$.
Within this formulation, VTAgent supports two core actions:

\quad $\cdot$ \textit{select key frame: $[F\_id_1,...]$}: This action identifies and extracts a subset of video frames that are most relevant to the given question by jointly considering visual content and visual text, and returns an ordered list of frame indices.

\quad $\cdot$ \textit{answer:} Once the keyframes are anchored, the model invokes this action to derive the final answer through text-centered reasoning based on the selected keyframes.

In summary, VTAgent formulates Video TextVQA as a structured agentic process that alternates between perception-driven decision making and targeted reasoning. This design enables MLLMs to suppress irrelevant visual redundancy and focus their reasoning on question-relevant evidence. Moreover, it inherently supports a training-free setting when the model possesses potent agentic abilities.

\begin{figure}[t!]
  \centering
    \includegraphics[width=\textwidth]{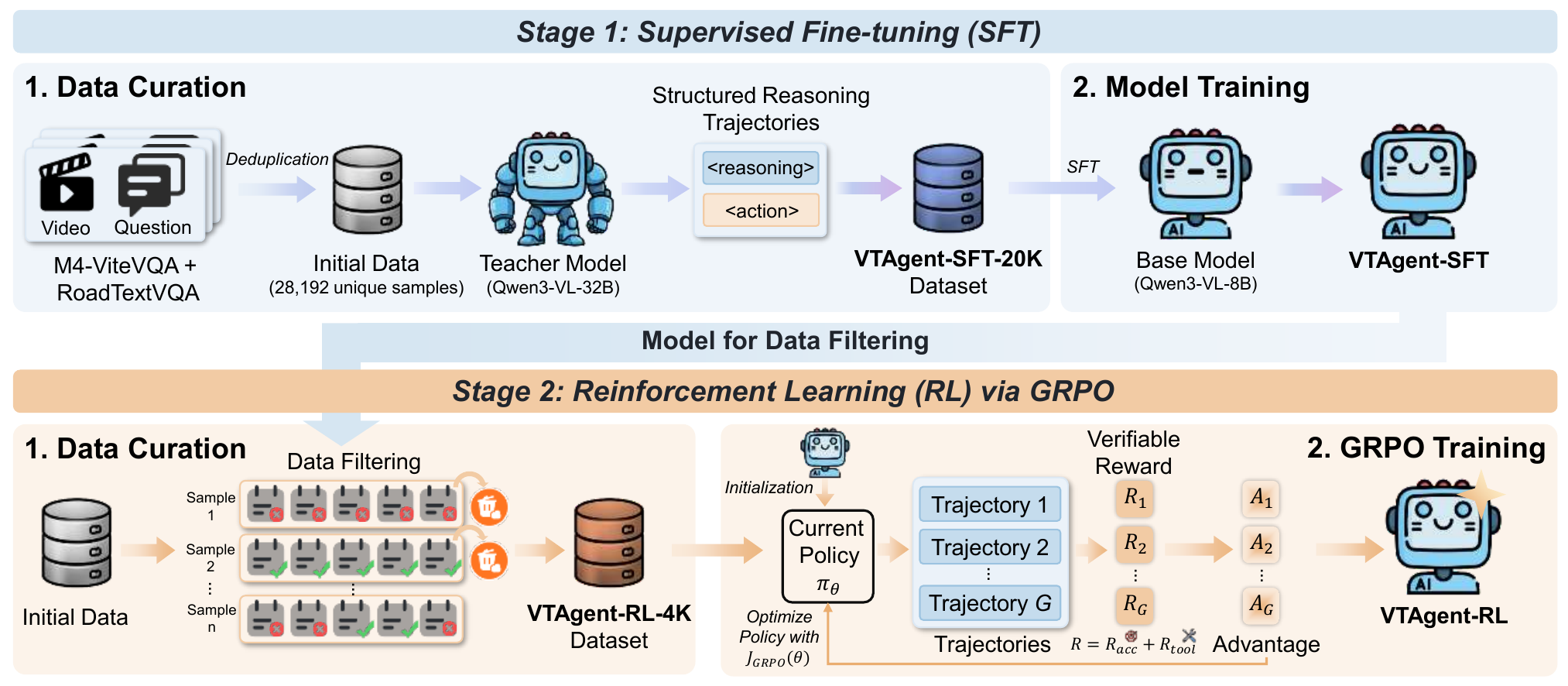}
    \caption{\textbf{Training pipeline of VTAgent.} }
    \vspace{-5pt}
\label{fig:train}
\end{figure}

\subsection{Supervised Fine-tuning (SFT)}
\label{sec:3.2}
Although recent MLLMs (\eg, Qwen3-VL~\cite{Qwen3-VL}) already exhibit strong agentic capabilities, their performance remains improvable through effective post-training. To this end, we introduce SFT to provide explicit and structured supervision, which guides the model to better align its intermediate reasoning trajectories with the corresponding agentic actions. As shown in Fig.~\ref{fig:train}, the SFT stage consists of two main components: (1) data curation, and (2) model training, which are described in detail below.

\noindent \textbf{Data Curation.} To construct high-quality supervision for agentic reasoning in Video TextVQA, we curate a dataset by leveraging existing open-source benchmarks and a strong teacher model. Specifically, we collect the training splits of two widely used Video TextVQA datasets, M4-ViteVQA~\cite{zhao2022towards} and RoadTextVQA~\cite{tom2023reading}, and perform deduplication to remove redundant samples. This process results in an initial dataset comprising 28,192 unique samples. We then employ a large multimodal model with strong agentic capabilities, Qwen3-VL-32B, to generate structured reasoning trajectories following the VTAgent framework. For samples where no valid answer is produced, the reasoning process is regenerated, with up to five attempts per sample. This iterative procedure ensures that each sample is paired with a valid and high-quality reasoning trajectory. As a result, we obtain the final SFT dataset, VTAgent-SFT-20K, consisting of 20,277 samples. For each sample, intermediate reasoning and executed actions are explicitly separated using the tags $<reasoning></reasoning>$ and $<action></action>$, providing clear structured supervision for agentic learning. The specific prompt templates used in this generation pipeline are provided in Appendix.

\noindent \textbf{Model Training.} Using the curated VTAgent-SFT-20K dataset, we adapt the base model (\eg, Qwen3-VL-8B~\cite{Qwen3-VL}) through SFT under a standard autoregressive language modeling paradigm. Each sample in VTAgent-SFT-20K is defined as a tuple:
\begin{equation}
D = (V, Q, T, Y) \in \mathcal{D}_{VTAgent-SFT-20K}, 
\end{equation}
where $V$ and $Q$ denote the input video and question, respectively, $T$ represents the structured agentic trajectory composed of interleaved reasoning and action segments, and $Y$ is the corresponding ground-truth answer. The model is trained to sequentially generate the complete trajectory $T$ and answer $Y$, which consist of interleaved $<reasoning>$ and $<action>$ sequences. Accordingly, the SFT objective is performed by minimizing the token-level cross-entropy over the target sequences. Through SFT, the model acquires an enhanced ability to identify question-relevant keyframes and to conduct coherent reasoning grounded in the selected visual evidence. Moreover, the resulting SFT model (\textbf{VTAgent-SFT}) provides a robust initialization for subsequent reinforcement learning, effectively stabilizing optimization, accelerating convergence, and maintaining the interpretability of intermediate reasoning traces.

\subsection{Reinforcement Learning (RL)}
\label{sec:3.3}
Although SFT effectively grounds agentic behaviors through explicit trajectory supervision, it remains limited by its reliance on static, offline annotations. In particular, SFT optimizes token-level imitation rather than task-level objectives, making it difficult to correct suboptimal reasoning paths or to balance accuracy and reasoning strategy~\cite{tan2025reason,long2025adsqa,deng2025openvlthinker}. RL offers a complementary paradigm by directly optimizing task-specific rewards, enabling adaptive refinement of reasoning and action policies beyond supervised trajectories. Motivated by these advantages, we employ Group Relative Policy Optimization (GRPO)~\cite{shao2024deepseekmath,guo2025deepseek} to further optimize the policy beyond the SFT stage.

\noindent \textbf{Data Curation.} To construct effective training data for RL, we leverage the model to focus optimization on appropriate challenging samples. Specifically, we employ VTAgent-SFT to answer each sample in the original initial data with 28,192 samples for up to five attempts. Samples for which the model consistently produces correct or error answers are excluded, as they offer limited informative feedback for model improvement. In contrast, samples that exhibit inconsistent outcomes across multiple attempts (\ie, a mixture of correct and incorrect predictions) are retained for RL training. This strategy allows the model to concentrate on failure-prone or ambiguous cases, where RL is most beneficial. Following this procedure, we obtain the VTAgent-RL-4K dataset, consisting of 4200 samples.

\noindent \textbf{Reward Design.} Given that VTAgent-SFT already exhibits strong compliance with the predefined output format, we omit format-related rewards in the reinforcement learning stage. Instead, the reward design focuses on task performance and effective tool utilization. To encourage the model to actively invoke the keyframe anchoring action, we introduce a tool usage reward that provides positive feedback when the model correctly executes the keyframe selection action. Consequently, the overall reward function consists of an answer correctness reward and a tool invocation reward, formulated as:
\begin{equation}
R = R_{acc} + R_{tool}, 
\end{equation}
where $R_{acc} \in \{0, 1\}$ indicates whether the predicted answer is correct, and $R_{tool} \in \{0, 0.5\}$ provides a positive incentive for invoking the keyframe selection action.

\noindent \textbf{GRPO Training.} For each sample $(V, Q, Y) \in \mathcal{D}_{VTAgent-RL-4K}$, we draw $G$ trajectories $\{\tau_i\}_{i=1}^G$ from the current policy $\pi_{\theta_{old}}$ and compute normalized outcome rewards to estimate relative advantages within each group. Following prior GRPO-based practices~\cite{yu2025dapo,zheng2025group,he2025framethinker}, we omit the KL regularization term to maintain sufficient flexibility during optimization, thereby avoiding overly restrictive policy updates that could impede effective exploration in the early stages of RL. The policy is optimized by maximizing:
\begin{equation}
\mathcal{J}_{GRPO}(\theta) = \mathbb{E}_{\tau_i \thicksim \pi_{\theta_{old}}}[\frac{1}{G} \sum_{i=1}^G \min(\frac{\pi_{\theta}}{\pi_{\theta_{old}}}A_i, \text{clip}(\frac{\pi_{\theta}}{\pi_{\theta_{old}}}, 1-\epsilon, 1+\epsilon)A_i)], 
\end{equation}
\begin{equation}
\text{where} \quad  A_i = \frac{R_i - \text{mean}(\{R_1,\dots,R_G\})}{\text{std}(\{R_1,\dots,R_G\}) + \delta}
\end{equation}
Here, $R_i$ denotes the reward associated with trajectory $\tau_i$, $\epsilon$ is the clipping hyperparameter, and $\delta$ is a small constant for numerical stability. Through the GRPO optimization, VTAgent is encouraged to refine its keyframe anchoring and reasoning behaviors based on relative outcome feedback, leading to more accurate answers.

\section{Experiment}

\subsection{Experimental Setup}

\noindent\textbf{Benchmarks.} We assess VTAgent on two widely used Video TextVQA benchmarks: \textbf{M4-ViteVQA}~\cite{zhao2022towards} and \textbf{RoadTextVQA}~\cite{tom2023reading}. \textbf{M4-ViteVQA} contains 8,511 video clips spanning nine scenario categories with three resolutions (720p, 1080p, and $1176 \times 664$), along with 24,123 QA pairs. It defines two tasks under three evaluation settings (Task1Split1, Task1Split2, and Task2) to assess standard performance, generalization, and domain adaptation. \textbf{RoadTextVQA} focuses on driver-assistance scenarios and consists of 3,222 driving videos paired with 10,500 QA instances. All videos have a resolution of $1280 \times 720$ at 30 FPS.

\noindent\textbf{Baseline Models.} To comprehensively evaluate the effectiveness of VTAgent, we compare it with two groups of baselines: (i) specialized methods tailored for Video TextVQA task, including T5-ViteVQA~\cite{zhao2022towards}, TEA~\cite{zhang2025track}, GAT~\cite{zhang2025gather}, ToM~\cite{yan2026tom}, and SFA~\cite{he2025sfa}; and (ii) general Video-LLMs, comprising Video-LLaVA~\cite{lin2024video}, VideoLLaMA2~\cite{cheng2024videollama}, NVILA~\cite{liu2025nvila}, Qwen2-VL-7B~\cite{wang2024qwen2}, InternVideo2.5-8B~\cite{zhang2025gather}, Qwen2.5-VL-7B~\cite{bai2025qwen2}, and Qwen3-VL-8B~\cite{Qwen3-VL}.

\begin{table*}[t!]
\centering
\caption{\textbf{Performance comparison of state-of-the-art methods on M4-ViteVQA Task1.} The best and second-best results are marked in \textbf{bold} and \underline{underlined}, respectively. \darkgreen{Green} numbers indicate the performance gains of VTAgent over the second-best method.}
\tiny
\setlength{\tabcolsep}{0.5pt}
\resizebox{\linewidth}{!}{{\begin{tabular}{lc|cccc|cccc}
\toprule[1pt]
\multirow{3}{*}{Methods} & \multirow{3}{*}{Year} &\multicolumn{4}{c|}{\textit{Task1Split1}} &\multicolumn{4}{c}{\textit{Task1Split2}}\\

\cline{3-10} & &\multicolumn{2}{c}{Validation} &\multicolumn{2}{c|}{Test} &\multicolumn{2}{c}{Validation} &\multicolumn{2}{c}{Test}\\

\cline{3-10} & &ACC. &ANLS &ACC. &ANLS &ACC. &ANLS &ACC. &ANLS\\
\hline

\rowcolor{gray!15} \multicolumn{10}{c}{\textbf{\textit{Specialized Methods}}} \\
T5-ViteVQA~\cite{zhao2022towards} &2022 &23.17 &30.10 &22.17 &29.10 &17.59 &23.10 &16.68 &23.80 \\

TEA-Base~\cite{zhang2025track} &2025 &34.45 &42.91 &31.70 &40.24 &26.66 &36.61 &26.29 &36.00 \\

TEA-Large~\cite{zhang2025track} &2025 &37.49 &46.38 &34.78 &43.71 &28.27 &36.32 &28.43 &38.13 \\

GAT-Base~\cite{zhang2025gather} &2025 &35.31 &44.64 &35.56 &45.21 &29.07 &39.26 &29.77 &40.71 \\

GAT-Large~\cite{zhang2025gather} &2025 &38.01 &47.53 &38.30 &48.23 &31.35 &41.33 &30.90 &41.81 \\

ToM~\cite{yan2026tom} &2026 &27.12 &35.5 &26.87 &35.4 &21.24 &28.71 &20.33 &27.5 \\

SFA~\cite{he2025sfa} &2025 &60.98 &68.62 &57.05 &65.44 &57.53 &66.21 &55.02 &\underline{64.63} \\

\hline

\rowcolor{gray!15} \multicolumn{10}{c}{\textbf{\textit{General Video-LLMs}}} \\
Video-LLaVA~\cite{lin2024video} &2024 &15.82 &17.77 &15.43 &17.15 &13.14 &14.29 &11.19 &12.02 \\

VideoLLaMA2~\cite{cheng2024videollama} &2024 &20.04 &21.73 &20.76 &23.55 &18.30 &19.63 &18.33 &20.45 \\

NVILA~\cite{liu2025nvila} &2025 &37.89 &47.67 &37.73 &47.23 &30.25 &40.58 &30.10 &41.52 \\

Qwen2-VL-7B~\cite{wang2024qwen2} &2024 &36.77 &46.56 &35.22 &45.84 &28.55 &39.34 &27.25 &38.45 \\

InternVideo2.5-8B~\cite{zhang2025gather} &2025 &39.83 &48.55 &40.0 &48.79 &41.60 &52.49 &38.99 &49.36 \\

Qwen2.5-VL-7B~\cite{bai2025qwen2} &2025 &58.35 &67.08 &56.11 &64.77 &54.69 &63.47 &50.93 &61.14 \\

Qwen3-VL-8B~\cite{Qwen3-VL} &2025 &\underline{61.95} &\underline{70.21} &\underline{57.96} &\underline{66.43} &\underline{58.70} &\underline{68.05} &\underline{55.35} &64.53 \\


Qwen3-VL-8B-Thinking~\cite{Qwen3-VL} &2025 &56.87 &65.99 &53.57 &62.73 &52.96 &63.63 &50.50 &60.84 \\

\hline

\multirow{2}{*}{VTAgent(ours)} &\multirow{2}{*}{-}  &\textbf{71.59} &\textbf{78.69} &\textbf{67.17} &\textbf{75.18} &\textbf{70.43} &\textbf{78.29} &\textbf{65.91} &\textbf{74.46} \\ 
& &(\darkgreen{$\uparrow$9.64}) &(\darkgreen{$\uparrow$8.48}) &(\darkgreen{$\uparrow$9.21}) &(\darkgreen{$\uparrow$8.75}) &(\darkgreen{$\uparrow$11.73}) &(\darkgreen{$\uparrow$10.24}) &(\darkgreen{$\uparrow$10.56}) &(\darkgreen{$\uparrow$9.83}) \\

\hline
\bottomrule[1pt]
\end{tabular}}}
\label{tab:1}
\vspace{-5mm}
\end{table*}

\noindent\textbf{Training Details.} Our implementation is built upon the Qwen3-VL-8B-Instruct model~\cite{Qwen3-VL} and trained on NVIDIA A800 GPUs. In the SFT stage, the model is fine-tuned on the VTAgent-SFT-20K dataset for 1 epoch using parameter-efficient LoRA~\cite{hu2022lora} with a rank of 8. We adopt an initial learning rate of $1 \times 10^{-5}$, with a cosine learning rate scheduler and a warmup ratio of 0.1. The batch size is set to 32. During the subsequent GRPO stage, the model is further optimized on the VTAgent-RL-4K dataset for 2 epochs with a batch size of 32. The learning rate is maintained at $1 \times 10^{-5}$, and the rollout number (i.e., the number of sampled trajectories per instance) is set to 4. Both training stages employ the same prompt template, as described in Appendix.

\subsection{Comparison with State-of-the-art Methods}
We evaluate VTAgent against existing state-of-the-art methods on the M4-ViteVQA and RoadTextVQA benchmarks, as detailed in Tab.~\ref{tab:1} and Tab.~\ref{tab:2}.
Across both datasets, VTAgent consistently outperforms all specialized methods and general Video-LLMs.
\textit{Notably, VTAgent surpasses the second-best performing method by an average margin of 12.12 in accuracy and 11.15 in ANLS}.

\noindent\textbf{Specialized Methods.} 
The integration of advanced OCR tools has enabled pioneering specialized methods~\cite{zhao2022towards,zhang2025track,zhang2025gather,yan2026tom} to achieve significant advancements in identifying scene text within videos.
Nevertheless, their broader video understanding capabilities remain constrained due to limitations in dataset scale and model parameter capacity, resulting in suboptimal performance and poor generalization.
For example, GAT~\cite{zhang2025gather} achieves accuracies of 38.30, 30.90, and 50.23 on the M4-ViteVQA Task1Split1
test set, Task1Split2 test set, and RoadTextVQA validation set, respectively.
However, GAT suffers a drastic performance drop under the cross-domain M4-ViteVQA Task2 setting, exposing its fragility in out-of-distribution scenarios
In contrast, SFA~\cite{he2025sfa} devises a training-free visual-cropping mechanism and apply it on Qwen2.5-VL-7B~\cite{bai2025qwen2}, yielding substantial gains over GAT.

\begin{table*}[t!]
\centering
\caption{\textbf{Performance comparison of state-of-the-art methods on M4-ViteVQA Task2 and RoadTextVQA.} The best and second-best results are marked in \textbf{bold} and \underline{underlined}, respectively. `Average' denotes the mean performance across the M4-ViteVQA and RoadTextVQA. \darkgreen{Green} numbers indicate the performance gains of VTAgent over the second-best method.}
\tiny
\setlength{\tabcolsep}{0.5pt}
\resizebox{\linewidth}{!}{{\begin{tabular}{lc|cccc|cc|cc}
\toprule[1pt]
\multirow{3}{*}{Methods} & \multirow{3}{*}{Year} &\multicolumn{4}{c|}{\textit{M4-ViteVQA Task2}} &\multicolumn{2}{c|}{\textit{RoadTextVQA}} &\multicolumn{2}{c}{\textit{Average}}\\

\cline{3-10} & &\multicolumn{2}{c}{Validation} &\multicolumn{2}{c|}{Test} &\multicolumn{2}{c|}{Validation} &\multicolumn{2}{c}{}\\

\cline{3-10} & &ACC. &ANLS &ACC. &ANLS &ACC. &ANLS &ACC. &ANLS\\
\hline

\rowcolor{gray!15} \multicolumn{10}{c}{\textbf{\textit{Specialized Methods}}} \\

T5-ViteVQA~\cite{zhao2022towards} &2022 &12.30 &16.10 &9.29 &13.60 &- &- &16.87 &22.63 \\

TEA-Base~\cite{zhang2025track} &2025 &20.73 &28.18 &17.28 &26.03 &44.43 &51.69 &28.79 &37.38 \\

TEA-Large~\cite{zhang2025track} &2025 &22.83 &30.21 &18.83 &28.99 &48.14 &54.85 &40.14 &48.31 \\

GAT-Base~\cite{zhang2025gather} &2025 &21.65 &30.88 &21.65 &29.83 &46.54 &53.78 &31.36 &40.62 \\

GAT-Large~\cite{zhang2025gather} &2025 &24.54 &33.30 &22.13 &30.75 &50.23 &58.12 &33.64 &43.01 \\

ToM~\cite{yan2026tom} &2026 &21.53 &25.2 &16.7 &22.4 &- &- &22.30 &29.12 \\

SFA~\cite{he2025sfa} &2025 &\underline{70.34} &\underline{76.60} &64.46 &71.72 &\underline{61.18} &\underline{67.28} &\underline{60.94} &68.64 \\

\hline

\rowcolor{gray!15} \multicolumn{10}{c}{\textbf{\textit{General Video-LLMs}}} \\

Video-LLaVA~\cite{lin2024video} &2024 &10.89 &13.23 &9.38 &11.80 &30.82 &40.92 &15.24 &18.17 \\

VideoLLaMA2~\cite{cheng2024videollama} &2024 &19.68 &23.62 &16.54 &21.80 &25.11 &36.53 &19.82 &23.90 \\

NVILA~\cite{liu2025nvila} &2025 &23.79 &32.89 &22.89 &30.34 &49.98 &57.22 &33.23 &42.49 \\

Qwen2-VL-7B~\cite{wang2024qwen2} &2024 &22.95 &32.65 &21.23 &28.79 &47.23 &55.34 &31.31 &41.00 \\

InternVideo2.5-8B~\cite{zhang2025gather} &2025 &48.03 &57.98 &41.36 &51.21 &41.96 &49.05 &41.68 &51.06 \\

Qwen2.5-VL-7B~\cite{bai2025qwen2} &2025 &66.40 &73.28 &62.98 &71.00 &50.33 &58.38 &57.11 &65.59 \\

Qwen3-VL-8B~\cite{Qwen3-VL} &2025 &69.82 &75.26 &\underline{66.50} &\underline{73.18} &53.38 &61.16 &60.52 &\underline{68.40} \\


Qwen3-VL-8B-Thinking~\cite{Qwen3-VL} &2025 &69.16 &75.79 &64.62 &72.02 &43.86 &53.09 &55.93 &65.01 \\
\hline

\multirow{2}{*}{VTAgent(ours)} &\multirow{2}{*}{-} &\textbf{80.45} &\textbf{84.86} &\textbf{76.90} &\textbf{82.72} &\textbf{78.97} &\textbf{84.36} &\textbf{73.06} &\textbf{79.79} \\ 
& &(\darkgreen{$\uparrow$10.11}) &(\darkgreen{$\uparrow$8.26}) &(\darkgreen{$\uparrow$10.4}) &(\darkgreen{$\uparrow$9.54}) &(\darkgreen{$\uparrow$17.79}) &(\darkgreen{$\uparrow$17.08}) &(\darkgreen{$\uparrow$12.12}) &(\darkgreen{$\uparrow$11.15}) \\

\hline
\bottomrule[1pt]
\end{tabular}}}
\label{tab:2}
\vspace{-5mm}
\end{table*}

\noindent\textbf{General Video-LLMs.}
In Tab.~\ref{tab:1} and Tab.~\ref{tab:2}, we have witnessed the rapid evolution of Video-LLMs in the domain of video comprehension. 
Early approaches, exemplified by Video-LLaVA~\cite{lin2024video}, VideoLLaMA2~\cite{cheng2024videollama}, and NVILA~\cite{liu2025nvila}, demonstrated limited performance on the Video TextVQA task, primarily attributed to inadequate video training corpora and a scarcity of OCR-specific data.
Subsequent advancements have obtained substantial performance gains, driven by the integration of large-scale OCR training corpora, enriched video collections, dedicated training strategies, and architectural innovations.
For instance, Qwen2.5-VL-7B and Qwen3-VL-8B~\cite{Qwen3-VL} attain an average of 57.11 and 60.52 accuracy across all evaluation datasets, respectively. 
Despite these strides, cutting-edge Video-LLMs remains constrained by the affection of redundant frames and imprecise localization of critical visual evidence.

\begin{figure}[t!]
  \centering
    \includegraphics[width=\textwidth]{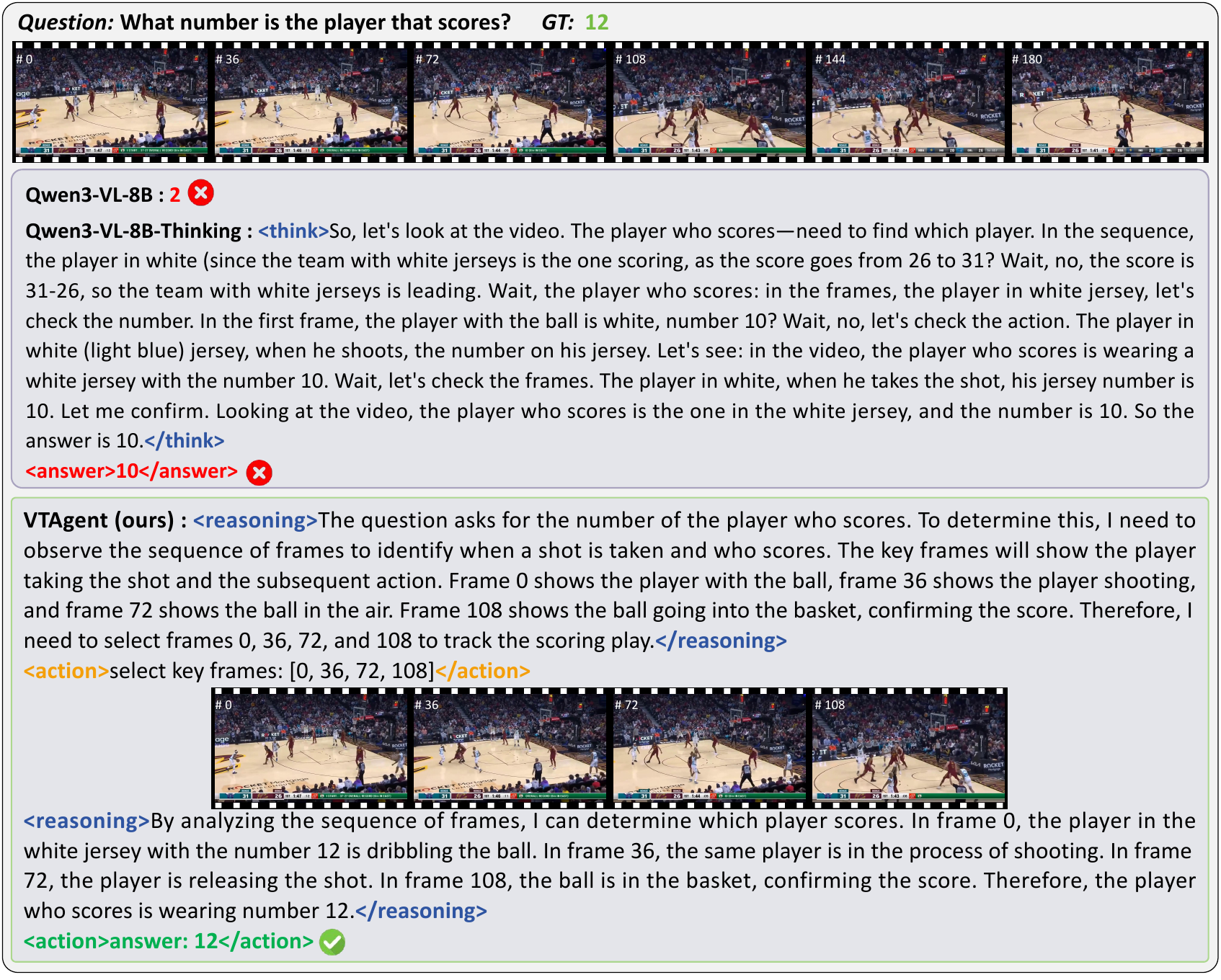}
    \vspace{-15pt}
    \caption{\textbf{A visualization example of VTAgent for keyframe anchoring and keyframe-conditioned reasoning.} }
    \vspace{-3mm}
\label{fig:vis}
\end{figure}

\noindent\textbf{VTAgent.}
As evidenced by the comprehensive results in Tab.~\ref{tab:1} and Tab.~\ref{tab:2}, VTAgent establishes a new state-of-the-art, demonstrating superior efficacy over all compared specialized approaches and general Video-LLMs.
In the M4-ViteVQA Task1 setting (Tab.~\ref{tab:1}), our method achieves an accuracy of 67.17 on the Split1 test set, surpassing the second-best competitor, Qwen3-VL-8B, by a significant margin of 9.21. 
This superiority is even more pronounced in the cross-domain and challenging settings presented in Tab.~\ref{tab:2}. Specifically, VTAgent attains 76.90 accuracy M4-ViteVQA Task2 test set, outperforming the runner-up by 10.4 accuracy. 
Notably, VTAgent exhibits remarkable robustness on the RoadTextVQA validation set with an accuracy of 78.97, exceeding the leading specialized method SFA by a substantial 17.79 points. 
These empirical results validate that VTAgent effectively addresses the critical bottlenecks of redundant frame interference and imprecise visual evidence localization inherent in prior works, thereby delivering exceptional comprehension for Video TextVQA.
Moreover, some qualitative visualizations are illustrated in Fig.~\ref{fig:vis}. VTAgent adopts a structured agent pipeline to locate the key frames, facilitating accurate answer generation. It begins with a global assessment on the question and video to localize critical frames, followed by a secondary and fine-grained reasoning stage focused exclusively on these selected key frames to produce final answer, mitigating the interference of redundant frames.

\subsection{Ablation Studies}

\noindent\textbf{Impact of Training Strategies.}
As reported in Tab.~\ref{tab:3}, applying the VTAgent pipeline on Qwen3-VL-8B in a training-free manner (VTAgent-TF) boosts average accuracy by 8.68 and improves average ANLS by 7.52. 
This underscores the intrinsic efficacy of our framework even without any parameter optimization. 
To further leverage stronger foundational representations, we scale the base model to Qwen3-VL-32B. This larger variant achieves superior baseline performance, motivating us to distill its high-quality reasoning trajectories into the more efficient 8B architecture.
After curating a high-quality subset from these distilled traces for the SFT stage, VTAgent-SFT attains 68.45 accuracy / 75.51 ANLS on average, establishing a robust initialization for the subsequent reinforcement learning stage. 
Finally, conducting reinforcement learning on the VTAgent-RL-4K dataset induces a pronounced performance leap, with additional gains of +6.83 in accuracy and +6.02 in ANLS.
In summary, these results validate that the staged training paradigm synergistically enhances the Video TextVQA comprehension.

\begin{table}[t!]
\centering
\caption{\textbf{Performance of VTAgent under different settings on validation sets of M4-ViteVQA Task1Split1 and RoadTextVQA.} `TF' denotes using VTAgent framework without training, while `SFT' and `RL' denote model trained with supervised fine-tuning and further reinforcement learning, respectively.}
\scriptsize
\setlength{\tabcolsep}{4pt}
\resizebox{\linewidth}{!}{{\begin{tabular}{l|c|cc|cc|cc}
\toprule[1pt]
\multirow{2}{*}{Methods} &\multirow{2}{*}{Base Model} &\multicolumn{2}{c|}{\textit{M4-ViteVQA T1S1}} &\multicolumn{2}{c|}{\textit{RoadTextVQA}} &\multicolumn{2}{c}{\textit{Average}}\\

\cline{3-8} & &ACC. &ANLS &ACC. &ANLS &ACC. &ANLS\\
\hline

Qwen3-VL-8B &- &61.95 &70.21 &53.38 &61.16 &57.67 &65.69 \\
VTAgent-TF &Qwen3-VL-8B &64.67 &72.08 &68.03 &74.33 &66.35 &73.21 \\
VTAgent-TF &Qwen3-VL-32B &66.11 &72.41 &69.65 &73.53 &67.88 &72.97 \\
VTAgent-SFT &Qwen3-VL-8B &66.21 &74.13 &70.69 &76.88 &68.45 &75.51 \\
VTAgent-RL &Qwen3-VL-8B &71.59 &78.69 &78.97 &84.36 &75.28 &81.53 \\

\hline
\bottomrule[1pt]
\end{tabular}}}
\label{tab:3}
\end{table}
\begin{table}[t!]
\centering
\caption{\textbf{Results of different reward function settings on validation sets of M4-ViteVQA Task1Split1 and RoadTextVQA.} \darkgreen{Green} numbers mean the improvements brought by the tool usage reward $R_{tool}$.}
\scriptsize
\setlength{\tabcolsep}{2pt}
\resizebox{\linewidth}{!}{{\begin{tabular}{cc|cc|cc|cc}
\toprule[1pt]
\multirow{2}{*}{$R_{acc}$} &\multirow{2}{*}{$R_{tool}$} &\multicolumn{2}{c|}{\textit{M4-ViteVQA T1S1}} &\multicolumn{2}{c|}{\textit{RoadTextVQA}} &\multicolumn{2}{c}{\textit{Average}}\\

\cline{3-8} & &ACC. &ANLS &ACC. &ANLS &ACC. &ANLS\\
\hline

$\checkmark$ &  &70.55 &77.59 &76.07 &81.18 &73.31 &79.39 \\
$\checkmark$ &$\checkmark$ &71.59(\darkgreen{$\uparrow$1.04}) &78.69(\darkgreen{$\uparrow$1.10}) &78.97(\darkgreen{$\uparrow$2.90}) &84.36(\darkgreen{$\uparrow$3.18}) &75.28(\darkgreen{$\uparrow$1.97}) &81.53(\darkgreen{$\uparrow$2.14}) \\

\hline
\bottomrule[1pt]
\end{tabular}}}
\vspace{-3mm}
\label{tab:4}
\end{table}

\noindent\textbf{Influence of Different Reward Functions.}
We conduct an ablation experiment to evaluate the influence of the tool invocation reward ($R_{tool}$) during RL training. Specifically, we compare training with only the answer correctness reward ($R_{acc}$) against training with both rewards. As shown in Tab.~\ref{tab:4}, using only $R_{acc}$ yields 70.55 accuracy / 77.59 ANLS on M4-ViteVQA and 76.07 accuracy / 81.18 ANLS on RoadTextVQA. By incorporating $R_{tool}$, the performance of VTAgent improves to 71.59 accuracy / 78.69 ANLS and 78.97 accuracy / 84.36 ANLS, respectively, resulting in average gains of $+1.97$ accuracy and $+2.14$ ANLS.
These results confirm that $R_{tool}$ contributes to more effective keyframe selection, which benefits downstream answering performance.

\subsection{Further Empirical Analysis of VTAgent}
To gain deeper insight into VTAgent, we conduct a targeted empirical study to disentangle its evidence localization and reasoning abilities. Guided by oracle frame-wise results, we divide the validation set of each M4-ViteVQA subtask into two groups: frame-solvable subset $Set_s$, where at least one individual frame independently yields the correct answer under frame-wise evaluation, and frame-unsolvable subset $Set_u$, where no single frame produces the correct prediction, indicating that successful answering requires multi-frame integration or more advanced reasoning. This partition enables a systematic evaluation of VTAgent under different capability requirements.

\begin{figure*}[t!]
\centering
  \subfloat[Keyframe hit rate of VTAgent on $Set_s$.]{\includegraphics[width=0.49\textwidth]{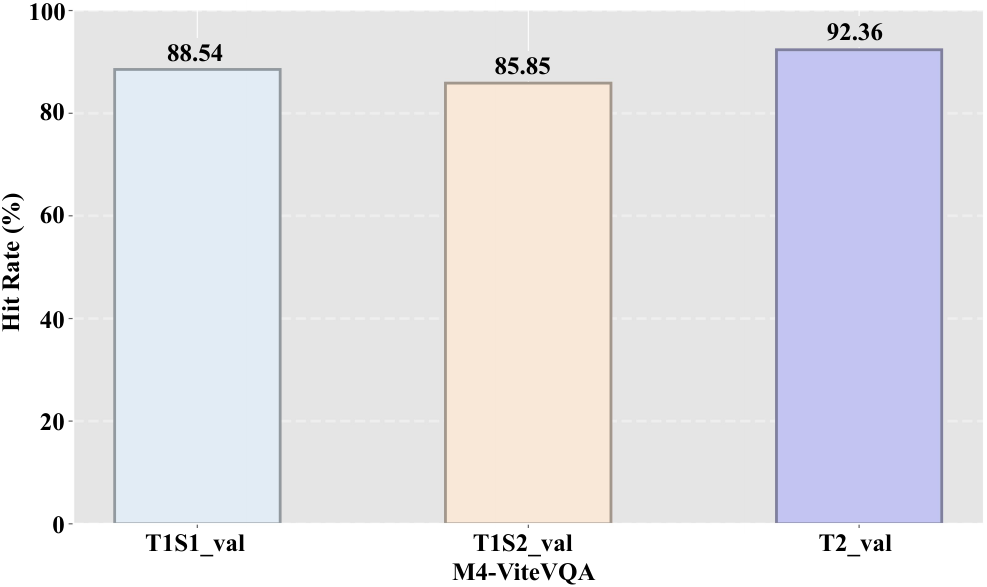}}
  \hfill
  \subfloat[Accuracy comparison on $Set_s$ and $Set_u$.]{\includegraphics[width=0.49\textwidth]{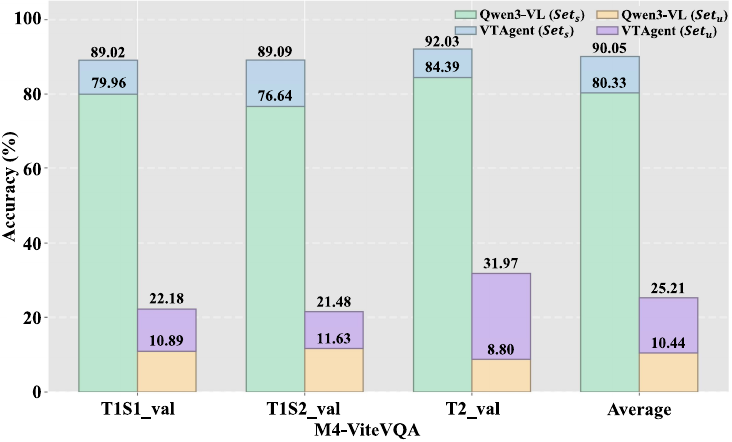}}
  \caption{\textbf{Analysis of VTAgent}. (a) demonstrates high keyframe hit rates, confirming reliable evidence localization; (b) shows superior answer accuracy over the baseline under varying task difficulties, highlighting effective reasoning with keyframe anchoring.}
  \vspace{-5pt}
 \label{fig:anly}
\end{figure*}

\noindent \textbf{Evidence Localization.} We first assess the evidence localization capability of VTAgent by measuring the accuracy of keyframe anchoring on the subset $Set_s$, where at least one frame can independently yield the correct answer under frame-wise inference and therefore naturally serves as a pseudo keyframe annotation. We adopt \textbf{Hit Rate} as the metric, defined as the proportion of samples for which at least one keyframe selected by VTAgent matches the annotated keyframe. As illustrated in Fig.~\ref{fig:anly} (a), VTAgent achieves hit rates of 88.54\%, 85.85\%, and 92.36\% on the three subsets of M4-ViteVQA, respectively. These consistently high scores provide direct evidence that VTAgent reliably identifies frames containing question-relevant textual evidence, effectively addressing the evidence localization bottleneck revealed in the oracle analysis and providing a solid foundation for subsequent reasoning.

\noindent \textbf{Reasoning.} We further compare the answer accuracy of VTAgent and the baseline Qwen3-VL-8B on both subsets to analyze their performance under different task difficulties. As shown in Fig.~\ref{fig:anly} (b), VTAgent consistently outperforms Qwen3-VL-8B on both subsets. On the frame-solvable subset $Set_s$, VTAgent achieves an average improvement of 9.72 points in answer accuracy over Qwen3-VL-8B. On the more challenging subset $Set_u$, which places higher demands on evidence localization and reasoning, the advantage further increases to 14.77 points. These results suggest that the explicit keyframe anchoring mechanism in VTAgent not only improves evidence grounding but also facilitates more reliable reasoning over video content, leading to consistent performance gains and better adaptability to more challenging scenarios.

\section{Conclusion}
In this work, we revisit the Video TextVQA task and identify a critical bottleneck in current Video-LLMs: the difficulty of accurately localizing question-relevant textual evidence amid temporally redundant and dynamically varying video content. 
Guided by this insight, we propose VTAgent, an agentic locate-and-focus framework that decomposes Video TextVQA into two sequential stages: keyframe anchoring and keyframe-conditioned reasoning. The framework operates effectively in a training-free setting and can be further enhanced through supervised fine-tuning and reinforcement learning, achieving new state-of-the-art results in both accuracy and ANLS across benchmarks.
Overall, we highlight the importance of explicit evidence localization in Video TextVQA and offer a practical and effective solution for evidence-grounded video reasoning.

\clearpage  


%
%
\bibliographystyle{splncs04}
\bibliography{main}
\end{document}